\documentclass[10pt,twocolumn,letterpaper]{article}

\usepackage[pagenumbers]{cvpr} 

\definecolor{cvprblue}{rgb}{0.21,0.49,0.74}
\usepackage[pagebackref,breaklinks,colorlinks,allcolors=cvprblue]{hyperref}
\usepackage{multirow}
\usepackage[accsupp]{axessibility}  
\def\paperID{7} 
\def\confName{CVPR}
\def\confYear{2026}

\title{Accuracy Improvement of Semi-Supervised Segmentation Using Supervised ClassMix and Sup-Unsup Feature Discriminator}

\author{
Takahiro Mano, 
Reiji Saito, and
Kazuhiro Hotta\\
Meijo University, 
1-501 Shiogamaguchi, Tempaku-ku, 
Nagoya 468-8502, Japan\\
{\tt\small \{180442126, 200442065\}@ccalumni.meijo-u.ac.jp, kazuhotta@meijo-u.ac.jp}
}

\begin{document}
\maketitle
\begin{abstract}
In semantic segmentation, the creation of pixel-level labels for training data incurs significant costs. To address this problem, semi-supervised learning, which utilizes a small number of labeled images alongside unlabeled images to enhance the performance, has gained attention. A conventional semi-supervised learning method, ClassMix, pastes class labels predicted from unlabeled images onto other images. However, since ClassMix performs operations using pseudo-labels obtained from unlabeled images, there is a risk of handling inaccurate labels. Additionally, there is a gap in data quality between labeled and unlabeled images, which can impact the feature maps. This study addresses these two issues. First, we propose a method where class labels from labeled images, along with the corresponding image regions, are pasted onto unlabeled images and their pseudo-labeled images. Second, we introduce a method that trains the model to make predictions on unlabeled images more similar to those on labeled images. Experiments on the Chase and COVID-19 datasets demonstrated an average improvement of 2.07$\%$ in mIoU compared to conventional semi-supervised learning methods.
\end{abstract}    
\section{Introduction}
\label{sec:introduction}

In recent years, with advancements in image recognition technology, various models have been proposed, such as the fully convolutional network FCN~~\cite{fcn}, encoder-decoder structures like SegNet~\cite{segnet} and U-Net~\cite{unet}, and the more advanced Deeplabv3+~\cite{deeplabv3plus}. However, a large amount of labeled data is generally required when performing image recognition using deep learning. Among these tasks, semantic segmentation is particularly demanding as it requires pixel-level labeling, making the preparation of a large dataset of labeled images costly.

In recent years, when using a large amount of labeled images, it has become possible to achieve high accuracy. However, when training with only a small number of labeled images, the accuracy significantly decreases. In real-world applications, it is desirable to reduce high costs by achieving high accuracy with only a limited number of labeled images. Against the background, a learning method called semi-supervised learning, which utilizes a small amount of labeled images alongside unlabeled images for model training, has garnered attention.

In semi-supervised segmentation, a technique called pseudo-labeling~\cite{pseudo} is the dominant approach. Semi-supervised segmentation heavily relies on the quality of these pseudo-labels. In medical imaging, which is the focus of this paper, there is often a significant class imbalance, making it difficult to predict rarely occurring classes, which in turn degrades the quality of pseudo-labels. Furthermore, since the model learns by treating the predicted pseudo-labels as ground truth, incorrect learning can lead to decreased accuracy. Therefore, when learning rare classes, it is crucial to ensure proper learning in the limited opportunities available.

A conventional semi-supervised segmentation method is ClassMix~\cite{classmix}. ClassMix involves cutting out the region of a randomly selected half of the predicted classes (e.g., one class if there are two) from one image and pasting it onto another image. By considering the shape of the class during the cut-and-paste process, ClassMix helps the model learn semantic boundaries between classes more effectively. However, ClassMix has two major issues. The first issue is that it performs ClassMix using predictions on unlabeled images. When ClassMix uses the prediction results for unlabeled images, the accuracy of the mixed images becomes dependent on the model’s prediction accuracy for unlabeled images. 
If the model makes incorrect predictions, the quality of the mixed images deteriorates. The second issue is that the regions of half the classes are selected randomly. Especially, since the background class has a large number of samples, it is likely to already have high accuracy. Therefore, learning by pasting classes that are already predicted with high accuracy does not provide much benefit.
Additionally, rare classes are less likely to become pseudo-labels because their prediction confidence remains low until learning progresses. This approach is not effective in datasets with significant class imbalances. To address these two issues, we propose a method called Supervised ClassMix (SupMix).

SupMix mixes regions except for background class, which has a large number of samples from labeled images, into pseudo-labels from unlabeled images. By attaching labels from different labeled images to the pseudo-labels, the accuracy of the pseudo-labels is improved without relying on the model's prediction accuracy, thereby addressing the issue of low accuracy in the initial pseudo-labels. Furthermore, by pasting regions other than background class from labeled images onto different unlabeled images, class imbalance can be mitigated. 
This helps to address the second issue of class imbalance.

In the research on semi-supervised learning, the domain gap between labeled and unlabeled images is often not considered. However, in real-world scenarios, there is an abundance of unlabeled images. If this domain shift can be properly addressed, semi-supervised learning can integrate more knowledge from unlabeled images. Therefore, this study focuses on minimizing the domain gap between predictions from labeled and unlabeled images. Specifically, we use a Generative Adversarial Network (GAN)~\cite{gan} to train the model so that the feature maps obtained from labeled images and those from unlabeled images are indistinguishable. By reducing the domain gap between the features extracted from labeled and unlabeled images, the model can efficiently acquire information from the unlabeled data, ultimately leading to improve the accuracy.

We conducted experiments on the Chase~\cite{Chase1,chase2} and COVID-19~\cite{covid19} datasets. Our goal was to improve the accuracy over UniMatch~\cite{unimatch}, a good precision method in semi-supervised segmentation. We compared the performance of UniMatch with our proposed method under conditions where only 1/4 and 1/8 of the total labeled images were used. Across all datasets, our proposed method achieved higher mIoU than the UniMatch.
For the Chase dataset, when we use 1/8 of the total labeled images, the IoU for the class with a small number of samples “retinal vessel” improved by 3.30$\%$ compared to UniMatch. When 1/4 of the labeled images is used, the IoU for “retinal vessel” increased by 2.63$\%$.
In the COVID-19 dataset, the IoU for the the class with a small number of samples “ground-glass” improved by 10.7$\%$ when we use 1/8 of the labeled images, and by 4.76$\%$  compared to UniMatch when 1/4 of the labeled images is used.

The structure of this paper is as follows: In Section \ref{sec:2_related}, we discuss related researches. Section \ref{sec:3_proposed} explains the details of the proposed method. In Section \ref{sec:4_experiments}, we present experimental results and provide a discussion. Finally, Section \ref{sec:5_conclusion} concludes the paper and outlines future challenges.
\section{Related Works}
\label{sec:2_related}

\subsection{Consistency regularization}

Consistency regularization frameworks~\cite{con1,con2,con3} are based on the idea that the predictions of unlabeled images should remain invariant even after applying augmentations. A common technique in classification tasks is called augmentation anchoring. Consistency regularization involves training in such a way that the predictions of augmented samples are forced to be consistent with the predictions the original unaugmented images. Our method utilizes this augmentation anchoring technique. The model is trained to maintain consistency between pseudo-labeled images, which are predictions of unlabeled images with weak augmentations, and synthetic images created by pasting the class shape of labeled images onto the pseudo-labeled images (SupMix). By using labeled images, the accuracy of the labels improves, ultimately leading to better overall performance.

\subsection{Augmentation methods}

The Cutout~\cite{cutout} algorithm is a technique that masks a square region within an image. By hiding specific partial areas, the model is encouraged not to rely on any particular region, allowing for a better understanding of the overall meaning of the image. The Random Erasing~\cite{randomerasing} algorithm, on the other hand, removes random rectangular areas. Unlike Cutout, it does not restrict itself to squares, and the size and location of the erased regions are determined randomly. The Mixup~\cite{mixup} algorithm is a method that linearly mixes two images and their corresponding labels, enabling the model to learn intermediate representations and become more robust to diverse data. The CutMix~\cite{cutmix} algorithm blends two different images by cutting out a random rectangular region from one image and pasting it onto another. CutMix also mixes the labels of both images based on the proportion of the rectangular area.

ClassMix~\cite{classmix} is an augmentation technique where randomly selected classes predicted from one image are cut out and pasted onto another image. Unlike CutMix, which cuts out a random rectangular region, leading to differences in context between the cut-out image and the destination image, making learning more difficult, ClassMix considers the shape of the class when cutting and pasting. This allows the model to learn the semantic boundaries of each class more effectively. However, conventional ClassMix relies on predictions from unlabeled images, which introduces the problem of depending on the prediction accuracy of those unlabeled images. To address this issue, we propose to paste a small number of classes from labeled images instead of relying on predictions from unlabeled images, which helps maintain the quality of pseudo-labels.

\subsection{Adversarial methods}

Generative Adversarial Networks (GANs)~\cite{gan} are used in various tasks beyond image generation, such as semantic segmentation~\cite{semgan,pix} and appearance inspection~\cite{anogan,effanogan,gannomal}, and have achieved strong performance. In the field of semi-supervised semantic segmentation, several methods leveraging GANs have also been proposed.

The first adversarial approach~\cite{first} used in semi-supervised semantic segmentation involves the generator increasing the number of samples available for training, while the discriminator also acts as the segmentation network. The output of the discriminator classifies each pixel as belonging to a correct class 
or a fake class. This enables the segmentation network to improve the ability to distinguish between real (supervised and unsupervised samples) and generated samples.
By treat supervised and unsupervised samples as the same class in discriminator, the method makes close the features of supervised and unsupervised samples indirectly.

However, none of the existing methods directly consider the domain gap between labeled and unlabeled images. Therefore, we aim to reduce the domain gap between the predictions of labeled and unlabeled images. To achieve this, we pass both labeled and unlabeled images through the model to obtain feature maps and then feed them into a discriminator, training it to distinguish between the two. The segmentation model is trained in such a way that it becomes difficult to differentiate whether the feature maps are from labeled or unlabeled images. This approach allows us to extract rich information from a large amount of unlabeled images, and we believe that it leads to improve the accuracy.

\section{Methodology}
\label{sec:3_proposed}

\begin{figure*}[t]
    \centering
    \includegraphics[width=1.0\linewidth]{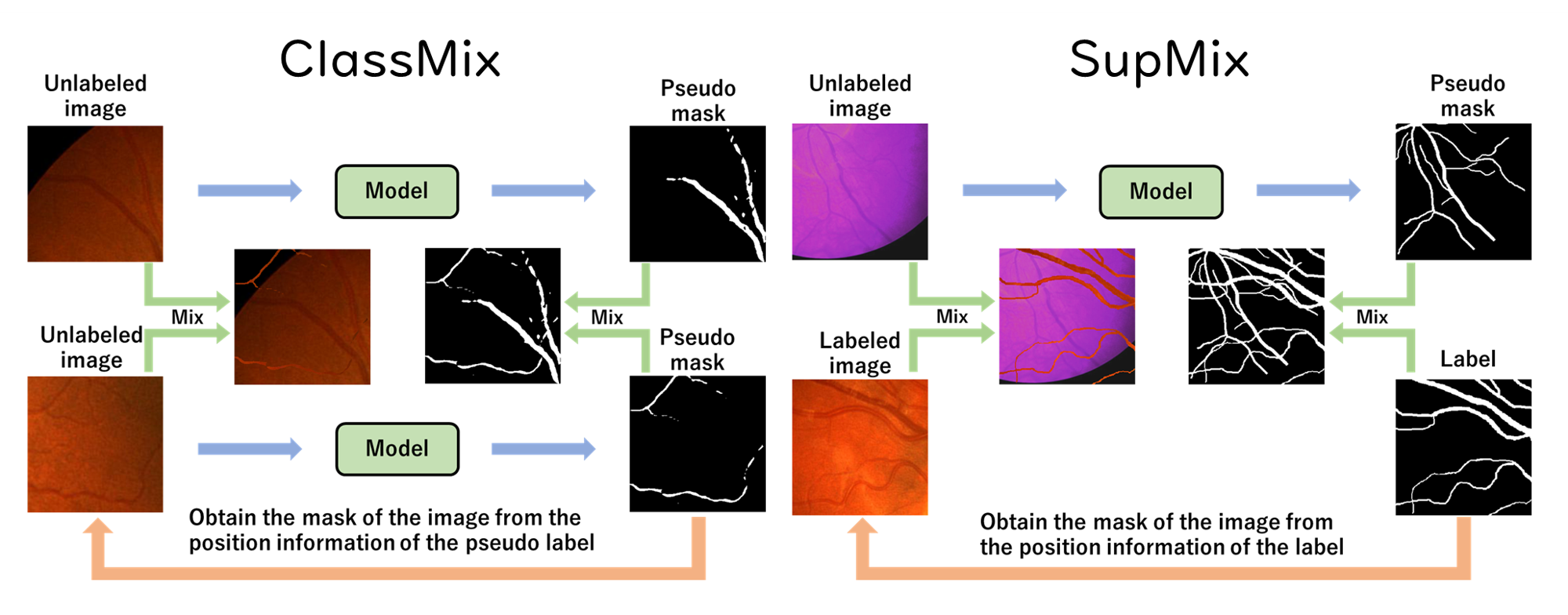}
    \caption{Comparison between the conventional ClassMix and the proposed SupMix. ClassMix is mixing images between images and their segmentation predictions.
    In contrast, SupMix pastes regions from a few classes in labeled images onto the pseudo-labels of unlabeled images. 
    The accuracy of the pseudo-labels mixed by SupMix is
    improved by using labeled images and pasting them onto unlabeled images with pseudo-labels. Since the ground-truth labels are independent of the prediction accuracy, only a few classes must be pasted onto the unlabeled images and pseudo-labels.}
    \label{fig:proposed}
\end{figure*}

We focus on semi-supervised segmentation, particularly with imbalanced medical datasets. We attempted two improvements. The first challenge is a modification of ClassMix~\cite{classmix} described in Section \ref{subsec:3_supmix}. The second challenge is to address the domain gap between the feature maps of labeled and unlabeled images. This is explained in Section\ref{subsec:3_disc}.

\subsection{Supervsed ClassMix(SupMix)}
\label{subsec:3_supmix}

We focus on enhancing ClassMix.
First, as a preliminary step for the subsequent improvements, we introduce ClassMix and clarify the issues. As shown on the left of Figure \ref{fig:proposed}, ClassMix is a technique where half of the predicted classes from one image are randomly selected, and their corresponding regions are cut out and pasted onto another image.
This method cuts and pastes regions while considering the shapes of the classes, allowing for more accurate learning of the semantic boundaries of each class. We will now introduce the ClassMix algorithm. First, we prepare two unlabeled images, $x_A\in \mathbb{R}^{3\times H \times W}$ and $x_B\in \mathbb{R}^{3\times H \times W}$, where $H$ and $W$ represent the height and width of the images. The two unlabeled images $x_A$ and $x_B$ are then fed into a model $f$ (such as DeepLabv3+~\cite{deeplabv3plus} specialized for segmentation). 
\begin{eqnarray}
  y_A &=& Argmax_c(f(x_A)) \\
  y_B &=& Argmax_c(f(x_B))
  \label{equation:pre_B}
\end{eqnarray}
where $y_A\in \mathbb{R}^{H \times W}$ and $y_B\in \mathbb{R}^{H \times W}$ represent the predicted class labels obtained by passing the input images through the model. Additionally, we retrieve the number of classes $\hat{C}$ present in $y_A$ and randomly select half of those classes. For an even number of classes (e.g., 2 classes), 1 class is selected. For an odd number of classes (e.g., 3 classes), half of the classes are selected by discarding the decimal part (e.g., 1 class). 
\begin{eqnarray}
  \hat{c} = \frac{\hat{C}}{2}
  \label{equation:c}
\end{eqnarray}

By using only the selected class $\hat{c}$ from the pseudo-label $y_A$ of the unlabeled input image $x_A$, a binary mask $M_A\in \mathbb{R}^{H \times W}$ is generated.
\begin{eqnarray}
\mathbf{M}_A = 
\begin{cases}
    1 & \text{if } \mathbf{y}_A \in \hat{c} \\
    0 & \text{otherwise}
\end{cases}
\label{equation:mask_A}
\end{eqnarray}
Finally, by using the binary mask, we generate the synthesized input image $x_{mix}\in \mathbb{R}^{3\times H \times W}$ and the synthesized pseudo-label $y_{mix}\in \mathbb{R}^{H \times W}$.
\begin{eqnarray}
  x_{mix} = M_A \odot x_A + (1 - M_A) \odot x_B
  \label{equation:mix_x}
\end{eqnarray}
\begin{eqnarray}
  y_{mix} = M_A \odot y_A + (1 - M_A) \odot y_B
  \label{equation:mix_y}
\end{eqnarray}
The synthesized input image $x_{mix}$ and the synthesized pseudo-label $y_{mix}$ represent the outputs of the ClassMix algorithm.

However, ClassMix has two issues. The first issue is that ClassMix is performed on both the unlabeled images and pseudo-labels. Since $y_A$ and $y_B$ equation \ref{equation:mix_y} are pseudo-labels obtained from unlabeled input images, the accuracy of $y_{mix}$ depends on the model's accuracy. As a result, if the model makes incorrect predictions, the accuracy of the mixed images will decrease (especially around the boundaries), making it difficult to improve the model's accuracy.

The second issue is that the classes for half the number of all classes are randomly selected and cut out for pasting. As shown in equation \ref{equation:c}, randomly selecting half number of classes can result in pasting regions from classes with high accuracy into another image, which reduces the effectiveness of training even when high-accuracy classes are used. Furthermore, since pseudo-labels are used to select half of the classes, classes with other than background class tend to progress more slowly in training and are often not included in the pseudo-labels. For these reasons, ClassMix cannot adequately handle datasets with significant class imbalance. To address these two issues, we propose Supervised ClassMix (SupMix).

We present the overview of SupMix on the right side of Figure \ref{fig:proposed}. Unlike ClassMix, which pastes pseudo-labels obtained from an unlabeled image onto another unlabeled image's pseudo-labels, SupMix mixes regions of specific classes obtained from a labeled image into the pseudo-labels of a weakly augmented unlabeled image. We define the labeled image and its label as $x_A^l$ and $y_A^l$, respectively. These are subject to weak preprocessing such as cropping and horizontal flipping. The key difference from ClassMix is the use of labeled data. Additionally, an unlabeled image $x_B^u$ is prepared, which is subjected to strong preprocessing (e.g., color jitter, blur, etc.). 
The unlabeled image $x_B^u$ is passed through the model $f$ to generate the pseudo-label $y_B^u$.
\begin{eqnarray}
  y_B^u &=& Argmax_c(f(x_B^u))
  \label{equation:pre_B_u}
\end{eqnarray}
Next, the number of existing classes $C^l$ is obtained from the ground truth labels. 
SupMix allows selecting classes to paste from all classes in the ground truth label, 
whereas ClassMix can only paste classes predicted within the pseudo-label
Background class is manually excluded, and we define this set as $C_{selected}$. The reason for manually excluding the background class compared to other classes is that it appears frequently during training due to its large number of samples, inducing its learning effectiveness. After excluding the background class, half of the remaining classes are randomly selected, which we define as $c^l_{selected}$. This serves as a solution to the second issue.
\begin{eqnarray}
  c^l_{selected} = \frac{C^l_{selected}}{2}
  \label{equation:c_t_se}
\end{eqnarray}

A binary mask $M_A \in \mathbb{R}^{H \times W}$ is generated from the pseudo-label $y_A^l$ of the unlabeled input image $x_A^l$, containing only the selected class $c^l{selected}$.
\begin{eqnarray}
\mathbf{M}_A^l = 
\begin{cases}
    1 & \text{if } \mathbf{y}_A^l \in c^l_{selected} \\
    0 & \text{otherwise}
\end{cases}
\label{equation:mask_A_t}
\end{eqnarray}
$M_A^l \in \mathbb{R}^{H \times W}$ is a binary mask obtained from the ground truth labels. This ensures that the accuracy of the mask is always maintained when pasting it onto another image. As a result, the pasting process does not rely on the model's accuracy, allowing it to be applied directly to the unlabeled image. This serves as a solution to the first issue and is the primary advantage of using SupMix.
Furthermore, with the modification of $M_A^l$, Equations \ref{equation:mix_x} and \ref{equation:mix_y} can be rewritten as follows.
\begin{eqnarray}
  x_{mix}^l = M_A^l \odot x_A^l + (1 - M_A^l) \odot x_B^u
  \label{equation:mix_x_l}
\end{eqnarray}
\begin{eqnarray}
  y_{mix}^l = M_A^l \odot y_A^l + (1 - M_A^l) \odot y_B^l
  \label{equation:mix_y_l}
\end{eqnarray}
The outputs $x_{mix}^l$ and $y_{mix}^l$ obtained from Equations \ref{equation:mix_x_l} and \ref{equation:mix_y_l} represent the final outputs of the SupMix algorithm. While there is a concern that pasting ground truth labels could lead to overfitting due to the repeated use of specific images or features, this is mitigated because the labeled images and their ground truth labels undergo preprocessing. Therefore, overfitting is considered less likely to occur.

\subsection{Sup-Unsup Feature Discriminator}
\label{subsec:3_disc}

In conventional semi-supervised learning, the domain gap between labeled and unlabeled images is not considered. However, in real-world scenarios, there is an abundance of unlabeled images. If this domain shift can be handled appropriately, semi-supervised learning can incorporate more knowledge from unlabeled images. Therefore, this paper proposes the Sup-Unsup Feature Discriminator (SUFD) to reduce the domain gap between the predictions of labeled and unlabeled images.

\begin{figure}[t]
\begin{center}
\includegraphics[scale=0.26]{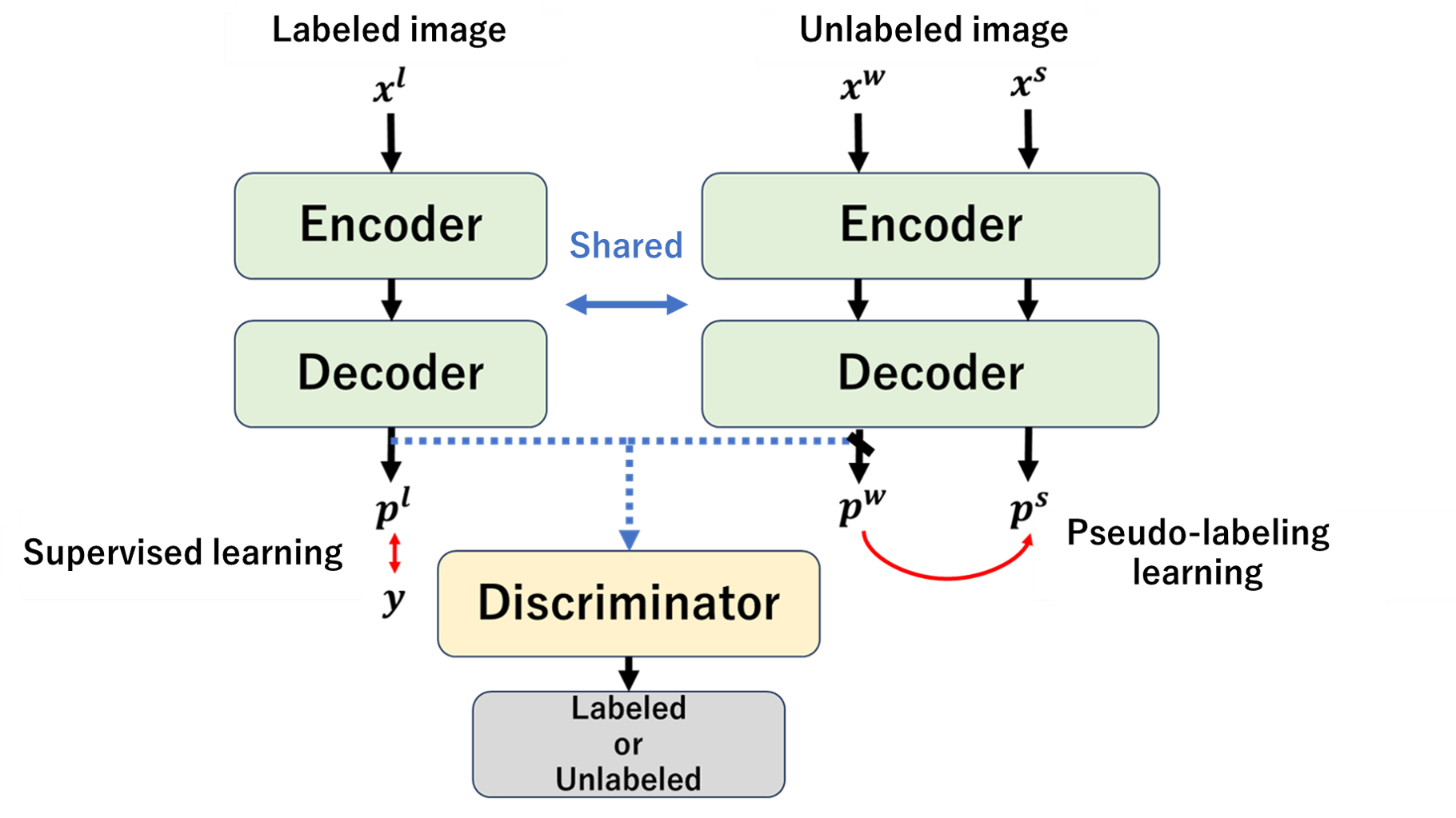}
\end{center}
\caption{The Sup-Unsup Feature Discriminator involves a shared encoder-decoder architecture used for both supervised learning and pseudo-supervised learning. Note that $x_l$ represents the labeled image, $x_w$ represents the unlabeled image (weak augmentation), and $x_s$ represents the unlabeled image (strong augmentation). $p_l$ denotes the output feature map of the labeled image, $p_w$ denotes the output feature map of the unlabeled image (weak augmentation), and $p_s$ denotes the output feature map of the unlabeled image (strong augmentation). Additionally, y represents the ground truth. During training, a discriminator is employed to make it difficult to distinguish between the supervised features and the unsupervised features without fixing, which are obtained after applying weak augmentations.}
\label{fig:SUFD}
\end{figure}

Figure \ref{fig:SUFD} provides the overview of the Sup-Unsup Feature Discriminator (SUFD). To reduce the domain gap between the predictions of labeled and unlabeled images, a discriminator commonly used in Generative Adversarial Networks (GAN), is employed. As shown in Figure \ref{fig:SUFD}, the discriminator is trained on the feature maps obtained from the model. The semi-supervised learning method, acting as a generator, is trained to produce feature maps from unlabeled images that the discriminator would mistake for features obtained from labeled images. In other words, this structure aims to make the feature maps derived from unlabeled images resemble features from labeled images, allowing the model to generate high-quality feature maps from unlabeled images similar to those from labeled ones.
Additionally, the loss functions used to train the generator and discriminator are provided as
\begin{eqnarray}
  \mathcal{L}_{SUFD}  =  B(o_u, 1) + \frac{1}{2}(B(o_u,0) + B(o_l,1)).
  \label{equation:SUFD}
\end{eqnarray}
where $o_u\in \mathbb{R}^{c\times h \times w}$ is the unsupervised feature map output by the discriminator, while $o_l\in \mathbb{R}^{c\times h \times w}$ represents the supervised feature map.
The first term is associated with learning the generator, while the second and third terms are related to training the discriminator. In the equation, B denotes the Binary Cross Entropy Loss. The discriminator performs a binary classification into two classes: 0 represents the “prediction from the unlabeled image”, and 1 represents the prediction from the labeled image”.

The generator is trained to make the discriminator misclassify unsupervised images as predictions from supervised images. This allows the feature maps obtained from unsupervised images to resemble those from supervised images. The discriminator is trained to output 0 when it identifies a prediction as being from an unsupervised image and 1 when it identifies it as being from a supervised image.
\section{Experiments}
\label{sec:4_experiments}

\subsection{Datasets and Implementation Details}
\label{sec:details}

The Chase dataset consists of a total of 28 images. The dataset size is $999 \times 960$, and for this experiment, we used 23 images for training and 5 images for testing. The task is to predict four classes: background and retinal vessels. The pixel ratio of each class in the ground truth labels was measured, with the background class at 93.36$\%$ and the retail vessel class at 6.64$\%$. From this, it is clear that there is a significant class imbalance.

The COVID-19 dataset consists of a total of 100 images, with 70 training images, 10 validation images, and 20 test images, all sized $256 \times 256$. The task is to predict four classes: background, ground-glass, consolidation, and pleural effusions. In this experiment, the validation images were not used; only the training and test images were utilized. When measuring the pixel ratio of each class in the ground truth labels, the background class is at 93.24$\%$, the ground-glass class is at 2.14$\%$, the consolidation class is at 4.50$\%$, and the pleural effusions class is at 0.12$\%$. From this, it is clear that the number of samples for classes other than the background class is significantly low.

We conduct experiments under conditions with limited supervision labels (semi-supervised learning). In this setting, we use 1/8 and 1/4 of the total training images as labeled images, while 7/8 and 3/4 are used as unlabeled images.
We compare supervised learning, FixMatch~\cite{fixmatch}, UniMatch~\cite{unimatch}, and the proposed method(ours). 
The model used in this experiment is Deeplabv3+~\cite{deeplabv3plus} with a ResNet-101 backbone~\cite{Resnet} pre-trained on ImageNet~\cite{imagenet}. 

The experiments were conducted using an Nvidia A6000. The batch size was set to 4, and the optimizer used was SGD with a momentum of 0.9 and a weight decay of $1 \times e^{-4}$. The initial learning rate for the scheduler was $4 \times e^{-3}$ for Chase and $1 \times e^{-3}$ for COVID-19, and it decayed after each iteration according to equation \ref{equation:lr}. The loss function used was Cross Entropy Loss. The model was trained for 1000 epochs on the Chase dataset and 500 epochs on the COVID-19 dataset.

Data augmentation in methods like FixMatch and UniMatch involves using weak and strong augmentations to learn from unlabeled images. The weak augmentations used include Random Crop and Random Horizontal Flip. The strong augmentations consist of these plus Random Color Jitter, Random Grayscale, Blur, and CutMix. The Random Crop is set to $320 \times 320$ for the Chase dataset and $256 \times 256$ for the COVID-19 dataset. The probability for Random horizontal Flip is set to 0.5. The probability for Random Color Jitter is 0.8, with brightness, contrast, and saturation all set to 0.5, and hue set to 0.25. The probability for Random Grayscale is 0.2, while the probabilities for Blur and CutMix are both set to 0.5.
All of these settings are the same as in UniMatch. In FixMatch and UniMatch, it is possible to set a threshold for pseudo labels, which is set at 0.95. Additionally, UniMatch employs dropout with a probability of 0.5. The evaluation metric was conducted using Intersection over Union (IoU), and the evaluation was based on the average of the experimental results obtained by changing the initializations five times.
\begin{eqnarray}
  lr_{current} = lr_{init} \times \left(\frac{1 - \text{iteration}}{\text{epoch}}\right)^{0.9}
  \label{equation:lr}
\end{eqnarray}

In the proposed method, instead of strong augmentation CutMix, Supervised ClassMix (SupMix) is used.
SupMix allows for specifying class labels when performing pasting. For both Chase and COVID-19, apart from the class with a large number of samples (background), half of the other classes (1 class for both Chase and COVID-19) are selected uniformly.

\subsection{Quantitative Evaluation}
\begin{table*}[h]
\caption{Accuracy on Chase and COVID-19 datasets for Supervised, FixMatch, UniMatch, and ours. The top table shows results on Chase and the bottom table shows the results on COVID-19 dataset. Each row shows the IoU and standard deviation for each class, while each column compares the case where the labeled data covers 1/8 (N images) and 1/4 (N images) in the total dataset. }
\label{table:Chase}
\centering
\scalebox{0.9}{
\begin{tabular}{c|cccc|cccc}
\hline
\multicolumn{1}{c|}{Chase} &
  \multicolumn{4}{c|}{1/8 (2 images)} &
  \multicolumn{4}{c}{1/4 (5 images)} \\ 
\hline
  
& Supervised & FixMatch & UniMatch & ours & Supervised & FixMatch & UniMatch & ours             \\
\hline
\hline

background & 74.62\tiny{$\pm$37.31} & 95.6\tiny{$\pm$0.12} & 96.41\tiny{$\pm$0.06} & \textcolor{red}{96.58\tiny{$\pm$0.06}} & 39.12\tiny{$\pm$44.57} & 96.48\tiny{$\pm$0.05} & 96.50\tiny{$\pm$0.02} & \textcolor{red}{96.63\tiny{$\pm$0.03}} \\

retinal vessel & 1.86\tiny{$\pm$2.54} & 38.54\tiny{$\pm$2.86} & 55.61\tiny{$\pm$0.81} & \textcolor{red}{58.91\tiny{$\pm$0.31}} & 3.78\tiny{$\pm$3.09} & 57.62\tiny{$\pm$0.60} & 57.78\tiny{$\pm$0.35} & \textcolor{red}{60.41\tiny{$\pm$0.20}} \\ 

mean IoU & 38.24\tiny{$\pm$17.51} & 67.07\tiny{$\pm$1.49} & 76.01\tiny{$\pm$0.43} & \textcolor{red}{77.74\tiny{$\pm$0.18}} & 21.45\tiny{$\pm$23.83} & 77.05\tiny{$\pm$0.32} & 77.14\tiny{$\pm$0.18} & \textcolor{red}{78.52\tiny{$\pm$0.10}} \\

\hline
\multicolumn{1}{c|}{COVID-19} &
  \multicolumn{4}{c|}{1/8 (9 images)} &
  \multicolumn{4}{c}{1/4 (18 images)} \\ 
\hline
\hline

background &  96.26\tiny{$\pm$0.07}& 95.57\tiny{$\pm$0.23} & 96.65\tiny{$\pm$0.13} & \textcolor{red}{96.72\tiny{$\pm$0.11}} & 96.78\tiny{$\pm$0.12} & 96.65\tiny{$\pm$0.08} & 97.03\tiny{$\pm$0.05} & \textcolor{red}{97.08\tiny{$\pm$0.07}}\\ 

ground-glass & 25.11\tiny{$\pm$0.97} & 33.52\tiny{$\pm$3.54} & 26.55\tiny{$\pm$3.94} & \textcolor{red}{37.25\tiny{$\pm$1.41}} & 32.50\tiny{$\pm$2.84} & \textcolor{red}{48.18\tiny{$\pm$1.06}} & 34.09\tiny{$\pm$1.21} &38.85\tiny{$\pm$1.85}\\ 
consolidation & 39.74\tiny{$\pm$2.32} & 0.0\tiny{$\pm$0.0} & 49.12\tiny{$\pm$1.65} & \textcolor{red}{50.86\tiny{$\pm$1.35}} & 45.12\tiny{$\pm$3.88} & 0.0\tiny{$\pm$0.0} & 47.57\tiny{$\pm$1.45} & \textcolor{red}{50.93\tiny{$\pm$1.14}}\\
pleural effusions & \textcolor{red}{0.0\tiny{$\pm$0.0}} & \textcolor{red}{0.0\tiny{$\pm$0.0}} & \textcolor{red}{0.0\tiny{$\pm$0.0}} & \textcolor{red}{0.0\tiny{$\pm$0.0}} & \textcolor{red}{0.0\tiny{$\pm$0.0}} & \textcolor{red}{0.0\tiny{$\pm$0.0}} & \textcolor{red}{0.0\tiny{$\pm$0.0}} & \textcolor{red}{0.0\tiny{$\pm$0.0}}\\
mean IoU & 40.28\tiny{$\pm$0.78} & 32.27\tiny{$\pm$0.87} & 43.08\tiny{$\pm$0.75} & \textcolor{red}{46.21\tiny{$\pm$0.60}} & 43.60\tiny{$\pm$1.51} & 36.21\tiny{$\pm$0.28} & 44.67\tiny{$\pm$0.63} & \textcolor{red}{46.71\tiny{$\pm$0.68}} \\

\hline
\end{tabular}
}
\end{table*}

The top Table \ref{table:Chase} shows the experiments on the Chase dataset. The proposed method (ours) outperformed UniMatch in both cases where the number of labeled images was 1/8 and 1/4. When we use 1/8 of the labeled images, our method achieved a 1.73$\%$ improvement in mIoU compared to UniMatch, with a notable 3.30$\%$ increase in accuracy for the retinal vessel class. When 1/4 of the labeled images is used, our method showed a 1.38$\%$ improvement in mIoU over UniMatch, with a 2.63$\%$ increase specifically for the retinal vessel class. The significant improvement in the retinal vessel class can be attributed to SupMix, which enhances learning opportunities for non-background areas by pasting pseudo-labels from different images. Additionally, the greater accuracy improvement with fewer labeled images is likely due to SUFD. SUFD is likely because the features of the unlabeled images effectively aligned with those of the labeled images, leading to the extraction of higher-quality features.

The bottom Table \ref{table:Chase} shows the experiments on the COVID-19 dataset. The proposed method (ours) outperformed UniMatch in both cases where the number of labeled images was 1/8 and 1/4. When we use 1/8 of the labeled images, our method achieved a 3.13$\%$ improvement in mIoU compared to UniMatch. Notably, the ground-glass and consolidation classes other than the background class, achieved accuracy improvements of 10.7$\%$ and 1.74$\%$. When 1/4 of the labeled images is used, our method showed a 2.04$\%$ improvement in mIoU over UniMatch. Notably, the ground-glass and consolidation classes other than the background class, achieved accuracy improvements of 4.74$\%$ and 3.36$\%$. The reason for improving the accuracy is the same as the discussion we made for the Chase dataset.

\subsection{Qualitative Evaluation}

\begin{figure*}[h]
    \centering
    \includegraphics[width=1.0\linewidth]{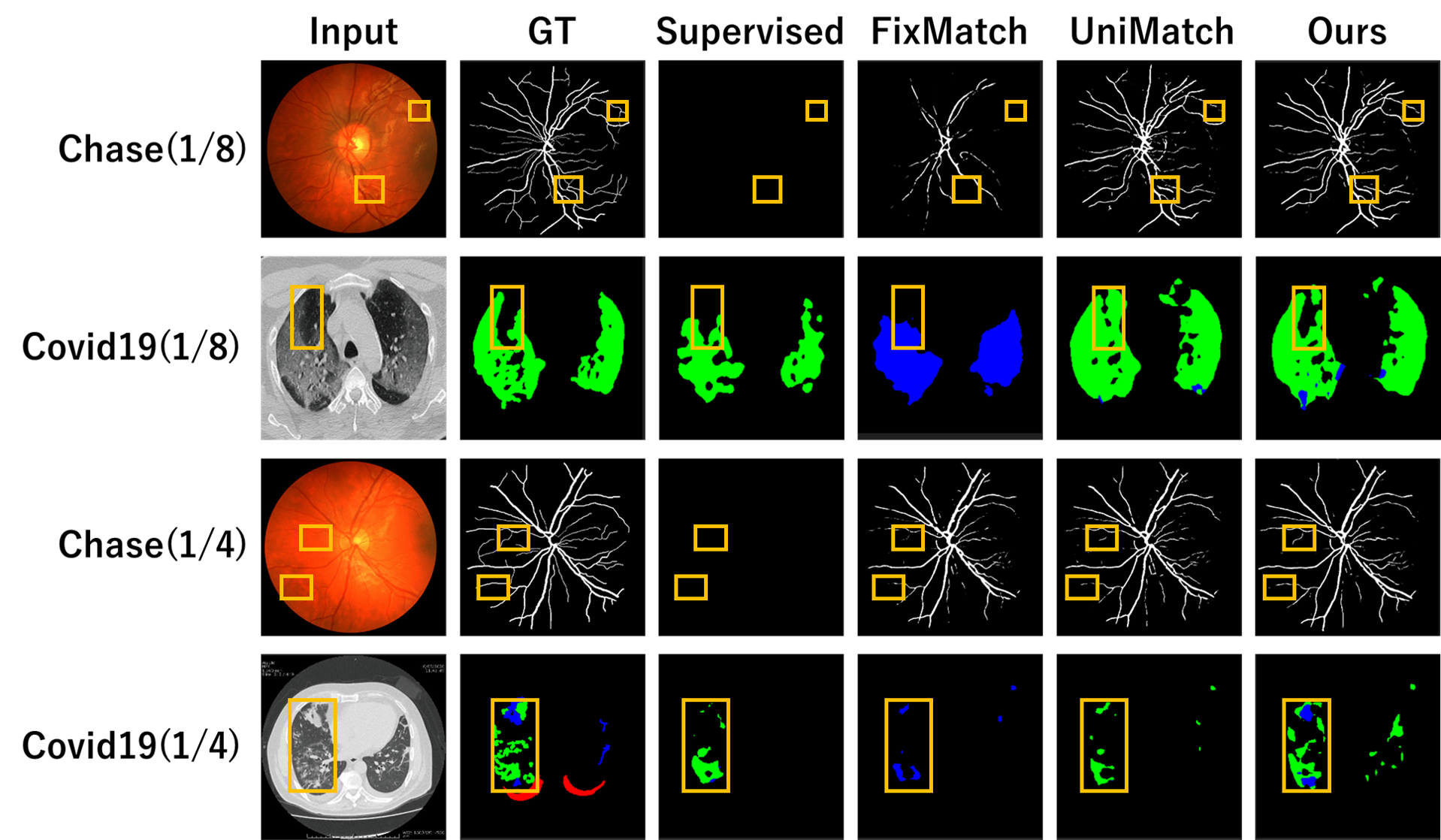}
    \caption{Segmentation results on Chase and COVID-19 datasets. Each row shows the dataset name and the ratio of supervised images used for training in all supervised images. Each column shows Input Images (Input), Ground Truth (GT), and the results of Supervised, FixMatch, UniMatch, and Ours. In the Chase dataset, the background class is visualized in black, and the retinal vessel class is in white. In the COVID-19 dataset, black represents the background class, blue indicates the ground-glass class, green shows the consolidation class, and red denotes the pleural effusions class.}
    \label{fig:covid_chase}
\end{figure*}

Figure \ref{fig:covid_chase} shows the segmentation results on the Chase and COVID-19 datasets. The areas highlighted in yellow indicate regions where accuracy has improved in comparison with UniMatch. In the Chase dataset, we see that the connectivity of retinal vessels has improved in the yellow areas for both 1/4 and 1/8 supervised images. This improvement is due to the SupMix method, where the supervised mask was pasted onto another image while maintaining the connectivity of the retinal vessels. Furthermore, the improvement in retinal vascular connectivity is larger for 1/8 than for 1/4 of all supervised images when we compare UniMatch with our method. This is because SUFD was able to gain more knowledge from the abundant unsupervised images.
Compared to UniMatch, for 1/8 of all supervised images in the COVID-19 dataset, the area within the yellow frame is predicted well in the background. For 1/4 of all supervised images, we see that the proposed method is closer to the ground truth than any other method within the yellow frame. These results demonstrate that our technique (SupMix and SUFD) is superior compared to other methods.

\subsection{Ablation Studies}

\begin{table*}[h]
\caption{Verification of the individual effects of SupMix and SUFD.
Each row indicates the dataset type and the classes with fewer samples (retinal vessel and ground-glass). In each column, the results display the accuracy
when we train the network with 1/N images in all supervised images. The values in parentheses are the improvement over CutMix.
The comparison methods include CutMix, ClassMix, SupMix, SUFD, and ours. CutMix corresponds to the standard UniMatch results, while “ours” refers to the combined method of SupMix and SUFD. 
}
\label{table:abu}
\centering
\scalebox{0.82}{
\begin{tabular}{c|ccccc|ccccc}
\hline
Methods & CutMix & ClassMix & SupMix & SUFD & ours & CutMix & ClassMix & SupMix & SUFD & ours              \\
\hline
\multicolumn{1}{c|}{Chase} &
  \multicolumn{5}{c|}{1/8 (2 images)} &
  \multicolumn{5}{c}{1/4 (5 images)} \\ 
  
\hline

retinal vessel & 55.61\tiny{\textcolor{red}{($+$0.00)}} & 56.78\tiny{\textcolor{red}{($+$1.17)}} & 57.57\tiny{\textcolor{red}{($+$1.96)}} & 55.95\tiny{\textcolor{red}{($+$0.34)}} & 58.91\tiny{\textcolor{red}{($+$3.30)}} &
57.78\tiny{\textcolor{red}{($+$0.00)}} & 57.91\tiny{\textcolor{red}{($+$0.13)}} & 59.60\tiny{\textcolor{red}{($+$1.82)}} & 56.90\tiny{\textcolor{blue}{($-$0.88)}} & 60.41\tiny{\textcolor{red}{($+$2.63)}}
\\

\hline
\multicolumn{1}{c|}{COVID-19} &
  \multicolumn{5}{c|}{1/8 (9 images)} &
  \multicolumn{5}{c}{1/4 (18 images)} \\ 
\hline

ground-glass &
26.55\tiny{\textcolor{red}{($+$0.00)}} & 27.95\tiny{\textcolor{red}{($+$1.40)}} & 32.82\tiny{\textcolor{red}{($+$6.27)}} & 33.36\tiny{\textcolor{red}{($+$6.81)}} & 37.25\tiny{\textcolor{red}{($+$10.70)}} &
34.09\tiny{\textcolor{red}{($+$0.00)}} & 34.99\tiny{\textcolor{red}{($+$0.90)}} & 38.10\tiny{\textcolor{red}{($+$4.01)}} & 34.17\tiny{\textcolor{red}{($+$0.07)}} & 38.85\tiny{\textcolor{red}{($+$4.76)}}
\\

\hline
\end{tabular}
}
\end{table*}

We introduced the SupMix and SUFD methods, but we have not yet verified the individual effectiveness of each method. Therefore, we conducted experiments to evaluate each method individually. The experimental procedures were carried out in the same manner as in Section \ref{sec:details}. The experimental results are shown in Table \ref{table:abu}. The baseline is UniMatch. Comparisons were made with conventional augmentation methods: CutMix and ClassMix.
SupMix and SUFD are our proposed methods, and the combination of these two is referred to as “ours” in the table. The experiments were conducted on the Chase and COVID-19 datasets, and results were obtained for cases where 1/8 and 1/4 of the fully supervised images were used. Additionally, only the accuracies for the classes with fewer samples (i.e., retinal vessels and ground-glass) are shown to verify if the methods address class imbalance effectively. 

First, regarding SupMix, it achieved significant accuracy improvements across all experimental results compared to conventional methods such as CutMix and ClassMix. This improvement is likely due to the fact that the class shapes from the labeled images were directly pasted onto other images, effectively addressing class imbalance. For SUFD, a notable accuracy improvement was observed when using 1/8 of the fully supervised images compared to conventional UniMatch (CutMix), whereas less improvement was seen when we use 1/4 of the images. This is because a large amount of information could be obtained from the abundant unlabeled images, leading to the observed accuracy gains. Finally, by combining these methods, we achieved the best results overall.
\section{Conclusion}
\label{sec:5_conclusion}

We propose a semi-supervised segmentation method using SupMix and SUFD, demonstrating superior results compared to conventional semi-supervised learning methods. However, since the accuracy for the most challenging pleural effusions class in the COVID-19 dataset did not improve, we plan to enhance the performance by considering prior probabilities and placing pleural effusions class labels in appropriate positions rather than simply pasting them. 
{
    \small
    \bibliographystyle{ieeenat_fullname}
    \bibliography{main}
}


\end{document}